\documentclass[10pt,twocolumn,letterpaper]{article}

\usepackage[pagenumbers]{cvpr} 

\usepackage{graphicx}
\usepackage{amsmath}
\usepackage{amssymb}
\usepackage{booktabs}
\usepackage{xcolor}
\usepackage{MnSymbol}

\usepackage[pagebackref,breaklinks,colorlinks]{hyperref}

\usepackage[capitalize]{cleveref}
\crefname{section}{Sec.}{Secs.}
\Crefname{section}{Section}{Sections}
\Crefname{table}{Table}{Tables}
\crefname{table}{Tab.}{Tabs.}

\newcommand{\firstcomponent}{cross-domain prediction consistency loss}
\newcommand{\secondcomponent}{cross-domain attention consistency loss}

\def\cvprPaperID{4767} 
\def\confName{CVPR}
\def\confYear{2023}

\begin{document}

\title{Exploring Consistency in Cross-Domain Transformer for Domain Adaptive Semantic Segmentation}


\author{Kaihong Wang$^1$ \and Donghyun Kim$^2$ \and Rogerio Feris$^2$ \and Kate Saenko$^{1,2}$ \and Margrit Betke$^1$ \\
$^1$ Boston University, $^2$MIT-IBM Watson AI Lab\\
{\tt\small \{kaiwkh, saenko, betke\}@bu.edu, dkim@ibm.com, rsferis@us.ibm.com}
}
\maketitle

\begin{abstract}

While transformers have greatly boosted performance in semantic segmentation, domain adaptive transformers are not yet well explored. We identify that the domain gap can cause discrepancies in self-attention. Due to this gap, the transformer attends to spurious regions or pixels, which deteriorates accuracy on the target domain. 
We propose to perform adaptation on attention maps with cross-domain attention layers that share features between the source and the target domains. Specifically, we impose consistency between predictions from cross-domain attention and self-attention modules to encourage similar distribution in the attention and output of the model across domains, i.e., attention-level and output-level alignment. We also enforce consistency in attention maps between different augmented views to further strengthen the attention-based alignment. Combining these two components, our method mitigates the discrepancy in attention maps across domains and further boosts the performance of the transformer under unsupervised domain adaptation settings. Our model outperforms the existing state-of-the-art baseline model on three widely used benchmarks, including GTAV-to-Cityscapes by 1.3 percent point (pp), Synthia-to-Cityscapes by 0.6 pp, and Cityscapes-to-ACDC by 1.1 pp, on average. Additionally, we verify the effectiveness and generalizability of our method through extensive experiments. Our code will be publicly available. 

\end{abstract}

\section{Introduction}
\label{sec:intro}

The transformer model has shown remarkable performance for various computer vision tasks and applications (\eg, \cite{DosovitskiyBeKoWeZhUnDeMiHeGeUsHo21,CarionMaSyUsKiZa20,XieWaYuAnAlLu21,YangQuNiYa21}) and often exhibits an outstanding prediction capacity compared to convolutional networks. Meanwhile, it is also relatively ``data-hungry'' and therefore expects a large amount of training data in order to achieve strong performance~\cite{TouvronCoDoMaSaJe21}. 
However, curating a large-scale annotated dataset could be a prohibitively expensive engineering task, 
especially for those problems that require pixel-level labeling, including semantic segmentation. Furthermore, deep models often generalize poorly to new domains such as different cities or  weather in driving scenes. To overcome these issues, the use of unsupervised domain adaptation (UDA) has been proposed. UDA  allows knowledge transfer from synthetic data (source domain), where pixel-level annotations are more cheaply available, to real-world data (unlabeled target domain). Recently, under the UDA setting for semantic segmentation, a transformer-based method (DAFormer~\cite{HoyerDaGo22}) outperforms previous CNN-based UDA methods across diverse benchmark datasets. 




\begin{figure}[t]
  \centering
   \includegraphics[width=\linewidth]{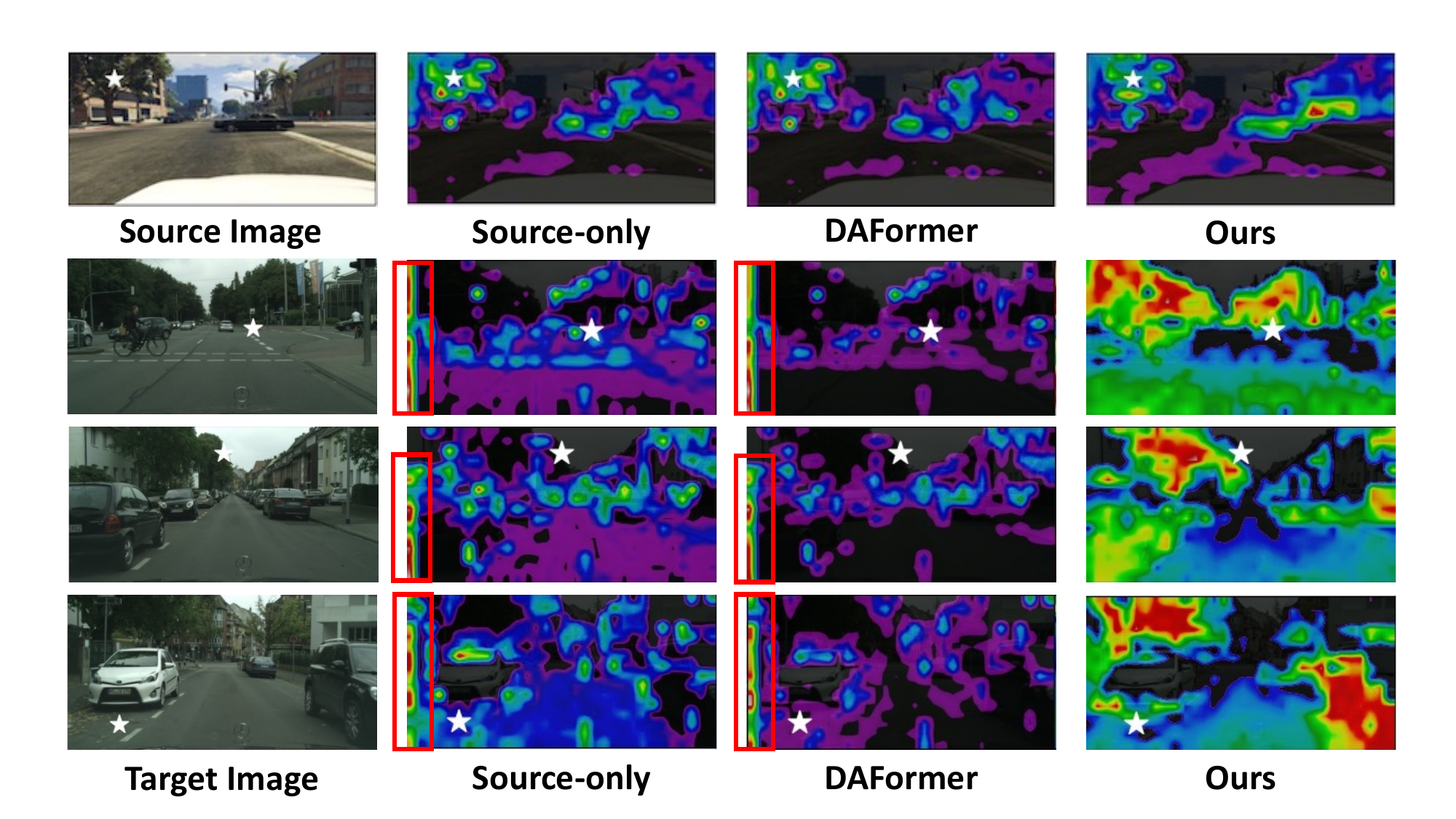}
   \caption{\small Attention map visualization for the query pixel $\largestar$ in source and target domains from different methods. The attention maps on the source image tend to highlight regions  sharing similar semantic classes. However, the attention maps on the target image from Source-only and prior work focus on spurious regions (\eg, left red boxes), which can be caused by domain gap.  In comparison, our method encourages attention-level adaptation and learns from more diverse and informative signals from both domains. 
}
  \label{fig:attn_teaser}
\end{figure}

The key component for the success in Transformer is \textit{self-attention}, which learns to attend to certain regions of its input that could be informative to predicting the semantic class of a pixel. However, it is still not clear if the domain gap is completely resolved under the UDA setting. We found that the attention maps from self-attention on the target images can focus on spurious regions and therefore fails to assist the prediction on the target domain as in Fig.~\ref{fig:attn_teaser}, which suggests that the domain gap still remains. We aim to improve the robustness of self-attention by using \textit{cross-domain attention}, which computes attention scores across different domain images. Previous work~\cite{XuChWaWaLiJi22} explores \textit{cross-domain attention} 
in image classification. However, this method requires a sophisticated search for finding positive cross-domain pairs, which may not be directly applicable to semantic segmentation. 

To more effectively mitigate the domain discrepancy and improve the robustness of Transformers for semantic segmentation, we propose our~\firstcomponent~to encourage consistency in the prediction (\ie, output-level) and attention map (\ie, attention-level). To be more specific, we at first compute predictions based on self-attention and cross-domain attention respectively. 
Then we supervised the predictions based on the self-attention module and the cross-domain attention module with the same label from either the source domain label or the target domain pseudo-label.
The benefits of this design are twofold: firstly, the consistency allows attention-level alignment
that helps pull the distributions of the self-attention and cross-domain attention closer as in Fig.~\ref{fig:attn_teaser}. Secondly, the operation also acts as a perturbation and regularization on the attention maps that encourage output-level domain alignment, \ie, facilitates consistent predictions when the attention maps differ due to the different input query vectors. In the end, the model benefits from both the alignment on attention-level between the self-attention and cross-domain attention as well as the more robust predictions on output-level.

Inspired by consistency learning \cite{TarvainenVa17,FrenchMaFi18} enforced on the output of models through different augmentations of an image, we propose our~\secondcomponent~to learn attention-level consistency and induce a model to generate robust attention maps. The idea is rather straightforward: The attention maps from two different augmented views from the same image should always be consistent.

With this objective, the model can learn to predict more consistent attention maps on both source and target images without supervision and further contribute to attention-level domain alignment. 

Combining our~\firstcomponent~and~\secondcomponent, our method not only facilitates alignment on output-level but also brings it to attention-level and improves the performance of the model under UDA settings. To summarize, our key contributions in this work include:

\begin{enumerate}
  \item We introduce \firstcomponent~that encourages robust attention as well as prediction across domains and therefore help the attention-level and output-level domain alignment;
  \item We propose to enforce attention-level consistency via our~\secondcomponent~to further assist the alignment of attention-level discrepancy;
  \item Our method is verified in extensive experiments to be effective, reliable and generalizable under different scenarios and achieves state-of-the-art performance on several important UDA semantic segmentation benchmarks including GTAV-to-Cityscapes, Synthia-to-Cityscapes and Cityscapes-to-ACDC.

\end{enumerate}

\section{Related Work}
\noindent\textbf{Unsupervised Domain Adaptation (UDA) for Semantic Segmentation.} UDA methods aim to leverage a labeled source domain dataset to learn on a target domain dataset without corresponding labels and reduce the domain shift. 
There are two major categories of existing methods in UDA semantic segmentation, the first one is to align domains into one distribution. Different approaches are proposed to achieve this goal including adversarial training~\cite{HoffmanWaYuDa16,HoffmanTzPaZhIsSaEfDa18,TsaiHuScSoHsCh18,VuJaBuCoPe19,TsaiSoScCh19,VuJaBuCoPe19,DuTaYaFeXuZhYeZh19,LuoZhGuYuYa19,ChoiKiKi19,WangYuWeFeXiHwHuSh20} on different levels including image, feature and output level, as well as non-adversarial approaches that such as style transfer~\cite{WuHLUGLD18,KimBy20,WangYaBe21}, Fourier transformation~\cite{YangSo20}.
Another category of methods focuses on the learning decision boundary on the target domain in an unsupervised manner, including self training~\cite{ZouYuKuWa18,ZouYuLiKuWa19,LiYuVa19,YangSo20,proda,TranhedenOlPiSv21} that learns from pseudo-labels, entropy minimization~\cite{VuJaBuCoPe19} and consistency regularization~\cite{ChoiKiKi19,WangYaBe21,AraslanovRo21}.
Recently, with DAFormer~\cite{HoyerDaGo22}, the Transformer model and adequate training strategies were introduced to the task of UDA semantic segmentation.  DAFormer greatly improves the state-of-the-art performance over the existing convnet based methods, while HRDA~\cite{HoyerDaGo22} further improves performance under high-resolution scenarios. We take DAFormer as our backbone and focus on attention-level alignment via cross-domain attention within Transformer rather than proposing a new architecture or training strategy.

\noindent\textbf{Transformers.} Inspired by transformer models used for NLP~\cite{vaswani2017attention}, vision transformers (\eg, ViT~\cite{dosovitskiy2021an}, DeiT~\cite{touvron2021training}, Swin~\cite{liu2021swin}, CSWin~\cite{DongBaChZhYuYuChGu22}) have been proposed that yield remarkable results for various computer vision tasks, including the task of semantic segmentation such as Segmenter~\cite{StrudelPiLaSc21}, SETR~\cite{ZhengLuZhZhLuWaFuFeXiToZh21}, and Segformer~\cite{XieWaYuAnAlLu21}. In multi-modal tasks (especially in language-vision), cross-modal attention between image and text has been widely exploited for aligning image and text representations (\eg,~\cite{albef,Chen2020UNITERUI,yu2022coca}).  

We aim to learn cross-attention across domains for knowledge transfer, where we do not have direct correspondences between source and target domains. The most relevant work is CDTrans~\cite{XuChWaWaLiJi22}, which shares features across domains in cross-domain attention for image classification. 
However, they require two-way labeling to find corresponding positives (\ie, the same category) in cross-attention, which is not directly applicable to the task of semantic segmentation. A concurrent work BCAT~\cite{WangGuZh22} concatenates features from both self-attention and cross-domain attention and uses the feature to perform domain alignment based on MMD~\cite{gretton2006kernel}.

In contrast to image classification, our method is proposed for semantic segmentation where multiple categories exist in a single image and therefore unsupervised learning for this task is more challenging. 
Our method does not require positive label pairing across domains but uses arbitrary (pseudo) labels to guide the consistent (output-level) prediction from the self-attention module as well as the noisy cross-domain attention module to improve robustness to the domain gap. 
In addition, we also address the attention-level discrepancy more directly with our~\secondcomponent.


\section{Method}
In this chapter, we first give a brief overview of the UDA setting and self-attention mechanism in Sec.~\ref{sec:preliminaries}, and then introduce our \firstcomponent~in Sec.~\ref{sec:cda_loss} and the attention-consistency loss in Sec.~\ref{sec:attn_loss}.

\subsection{Preliminaries} \label{sec:preliminaries}
Formally, given a source domain dataset $D_s=\{(x_s^i, y_s^i)\}_{i=1}^{N^s}$ with corresponding labels and an unlabeled target domain dataset $D_t=\{x_t^i\}_{i=1}^{N^t}$, our task is to learn a student semantic segmentation model $F$ with its self-ensemble teacher $G$ and optimize the student’s performance on the target domain. At step $i$, the teacher’s weights $\phi_i$ are updated as the moving average of the student's weight $\theta_i$ as $\phi_i = \alpha \phi_{i-1}+(1-\alpha)\theta_{i}$. 

In a multi-head self-attention module, an input feature map at the dimension of $N \times d$, where $N = H \times W$, will be passed through a query weight $W^Q$, a key weight $W^K$ and a value weight $W^V$ with the shape of $d \times C$ to produce a query vector $Q$, a key vector $K$, and value vector $V$ sharing the same dimensions of $N \times C$. Thus, the self-attention can be obtained by:
\begin{equation}
    \text{Attention}(Q,K,V)= \text{Softmax}(\frac{QK^T}{\sqrt{d_{head}}})V,
  \label{eq:self-attn}
\end{equation}
where $d_{head}$ is the number of dimensions of one head.

Typical UDA segmentation pipelines, including DAFormer~\cite{HoyerDaGo22}, consist of a supervised branch for the source domain and an unsupervised branch for the target domain. The supervised learning branch computes the cross-entropy loss $L_s$ between the prediction of the student $F$ on a source image $x_s$ and its corresponding label $y_s$:
\begin{equation}
    L_s^{(i)} = -\sum_{j=1}^{H\times W} \sum_{c=1}^C y_s^{(i,j,c)} \log F(x_s^i)^{(j,c)},
  \label{eq:sup_v1}
\end{equation}
where $H$ and $W$ are the height and width of an input image, respectively, while $C$ is the number of categories shared between the source and the target domain. 
Following DACS~\cite{TranhedenOlPiSv21}, a target image $x_t$ and its pseudo-label $p_t$ predicted by $G$ (teacher) will be mixed with $x_s$ and $y_s$ through a binary mixing mask $M$ in the same shape with $x_s$.
Then mask $M$ is applied to obtain the augmented image $\hat{x}_t$ and label $\hat{p}_t$ for the target domain supervision. If a pixel at $j$ from the mask $M^j=1$, then $\hat{x}_t^{i,j}=x_s^{i,j}$ and $\hat{p}_t^{i,j}=y_s^{i,j}$, otherwise $\hat{x}_t^{i,j}=x_t^{i,j}$ and $\hat{p}_t^{i,j}=p_t^{i,j}$. This yields the loss for the unsupervised branch:
\begin{equation}
L_t^{(i)} = -\sum_{j=1}^{H\times W} \sum_{c=1}^{C} q^i \hat{p}{_t}^{(i,j,c)} \log F(\hat{x}_t^i)^{(j,c)},
  \label{eq:unsup_v1}
\end{equation}
where $q^i$ is the weight mask for $\hat{x}_t^i$ as in DAFormer. We take DAFormer as our backbone model.

\begin{figure}[t]
  \centering
   \includegraphics[width=\linewidth]{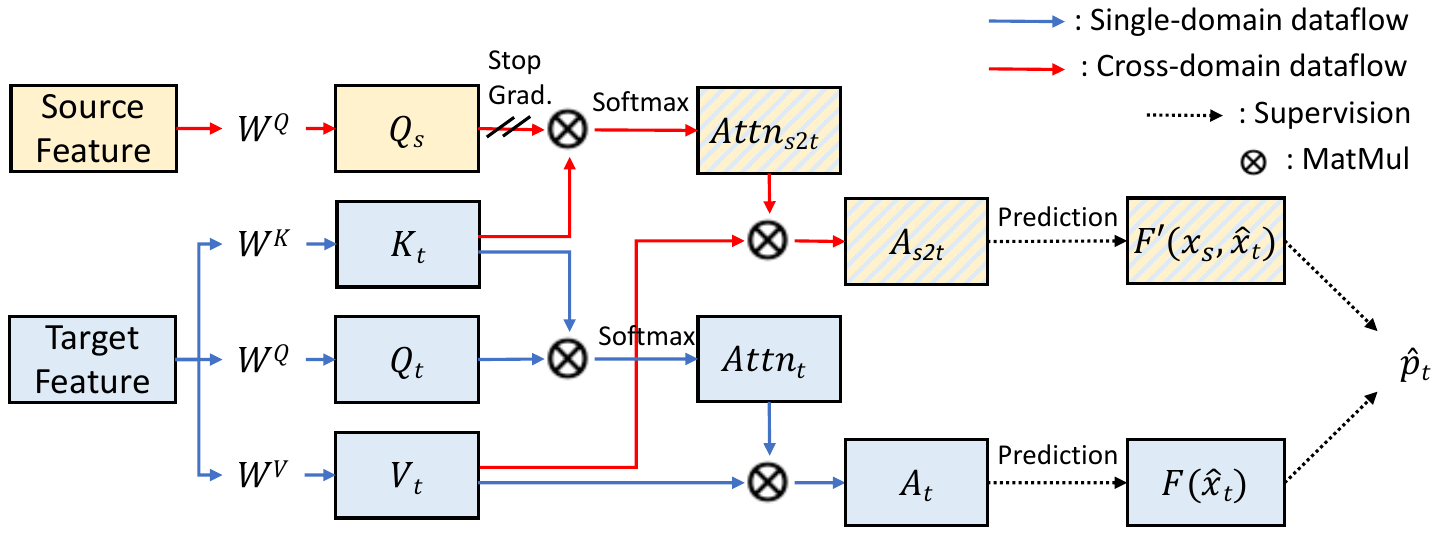}

   \caption{\small Illustration of our consistency learning with Cross-domain Attention module for UDA semantic segmentation on the target and source-to-target branches. We swap the query vectors so the cross-domain branches take features from both domains. The computation for the source and target-to-source branches is symmetrical, except that the supervision here is the pseudo-label $\hat{p}_t$ generated from the target image $\hat{x}_t$ while that for the source and target-to-source branches is from the source label $y_s$.
}
   \label{fig:cda}
\end{figure}

\subsection{Cross-Domain Prediction Consistency Loss} \label{sec:cda_loss}
While existing Transformers rely on self-attention that attends regions in the same image that will the informative, we found that the attention often could be spurious and focus on non-informative regions. This becomes more serious when there are distributional shifts in test data as shown in Fig.~\ref{fig:attn_teaser}. Some of the prior works~\cite{NamKiHeHaOhOh2,WuGaZhLiXiSuLi22} are proposed to improve self-attention but their methods do not work well in our adaptation studies. 

Since a model is mostly supervised with source labels, it can lead to overfitting to source domains. For the target domain from different distributions, self-attention can be noisy and attend to less informative regions. We compute the prediction based on the cross-domain attention and enforce its consistency to that of the self-attention. The enforcement of consistency not only helps bridging distributional shifts in attention across domains, but also leads a model to be robust even when the attention map is noisy. This can also be understood as regularization of the self-attention mechanism with our cross-domain attention layers in a similar spirit to other regularization methods such as Dropout~\cite{SrivastavaHiKrSuSa14}. 
In Sec.~\ref{sec:pert}, we also compare our method with other regularization approaches.

Specifically, through the same query weight $W^Q$, key weight $W^K$ and value weight $W^V$, we extract source domain vectors $Q_s$, $K_s$, and $V_s$ from source image $x_s$ and $Q_t$, $K_t$, and $V_t$ from target image $x_t$. 
In opposition to the typical self-attention module in a regular model $F$ that takes these vectors only from the same image as $A_{s}=\text{Attention}(Q_s,K_s,V_s)$ and $A_{t}=\text{Attention}(Q_t,K_t,V_t)$, we swap the query vectors to produce cross-domain attention. A cross-domain model $F'$ sharing the same architecture and weights with the student $F$ is applied to take a pair of images from different domains.
$F'(\hat{x}_t,x_s)$ extracts initial features from the source image $x_s$ and produces output based on the target-to-source attention $A_{t2s}=\text{Attention}(Q_t,K_s,V_s)$. Symmetrically, $F'(x_s,\hat{x}_t)$ extracts initial features from the target image $\hat{x}_t$ and computes the output based on the source-to-target attention $A_{s2t}=\text{Attention}(Q_s,K_t,V_t)$. 
Since we consider the attention from different query feature vectors in the other domain can be noisy, we use gradient stopping on the query features vectors from the other domain. 
The target-to-source prediction will be guided by the source label (Eq.~\ref{eq:sup_v1}) while the source-to-target prediction will be guided by the pseudo-label from $G$ (Eq.~\ref{eq:unsup_v1}), yielding our proposed losses:
\begin{equation}  \label{eq:cda_losses}
    \begin{split}
    L_{t2s}^{(i)} = -\sum_{j=1}^{H\times W} \sum_{c=1}^C y_s^{(i,j,c)} \log F'(\hat{x}_t^i,x_s^i)^{(j,c)} \\
    L_{s2t}^{(i)} = -\sum_{j=1}^{H\times W} \sum_{c=1}^{C} q^i \hat{p}{_t}^{(i,j,c)} \log F'(x_s^i,\hat{x}_t^i)^{(j,c)}.
  \end{split}
\end{equation}
Finally, we take the average of the losses from the source and target supervision and construct our \firstcomponent~into $L_{pred}$: 
\begin{equation}  \label{eq:cda_loss_final}
    \begin{split}
    L_{pred}=\frac{1}{2}(L_{s} + L_{t2s}) + \frac{1}{2}(L_{t} + L_{s2t}).
    \end{split}
\end{equation}




\begin{figure}[t]
  \centering
   \includegraphics[width=\linewidth]{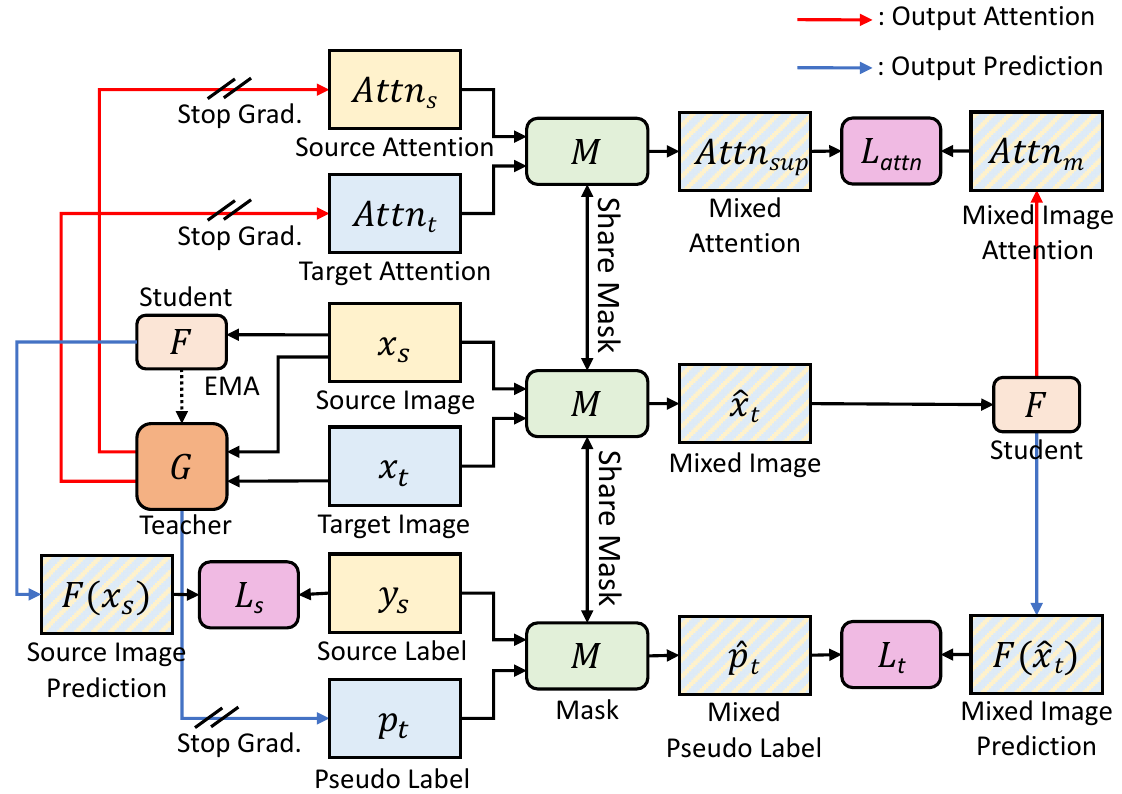}

   \caption{\small Illustration of our~\secondcomponent~and its relation to $L_{s}$ and $L_{t}$. With the mean-teacher framework that guides the training of the student model by the prediction of the teacher on an unlabeled target image through augmentations (DACS), we propose to enforce the consistency on the intermediate attention maps in a similar way. 
}
   \label{fig:attn_losses}
\end{figure}

\subsection{Cross-Domain Attention Consistency Loss} \label{sec:attn_loss}
We further regularize a model by consistency on attention maps. We enforce the model to produce consistent attention maps from different augmentations to strengthen the alignment on attention-level. Fig.~\ref{fig:attn_losses} illustrate the overview of our~\secondcomponent. Firstly, we extract attention maps $\{Attn_{s}^i\}_{i=1}^L$ and $\{Attn_{t}^i\}_{i=1}^L$ via $L$ attention modules of the teacher model $G$ from the source image $x_s$ and the target image $x_t$, as well as $\{Attn_{m}^i\}_{i=1}^L$ via the student model $F$ from the mixed image $\hat{x}_t$ (in Sec.~\ref{sec:preliminaries}).
The source and the target attention maps $\{Attn_{s}^i\}_{i=1}^L$ and $\{Attn_{t}^i\}_{i=1}^L$ will be mixed by $M$ that produces $\hat{x}_t$.  Then we compute the mixed attention maps $\{Attn_{sup}^i\}_{i=1}^L$ to regularize the consistency on attention maps
\begin{equation}
  \label{eq:mix_attention_maps}
Attn_{sup}^i[:,:,j,:] =\left\{
\begin{aligned}
Attn_{s}^i[:,:,j,:]  & , & M'^j=1, \\
Attn_{t}^i[:,:,j,:]  & , & M'^j=0,
\end{aligned}
\right.
\end{equation}
where $M'$ is resized from $M$  and j represents the spatial location.
Finally, we minimize the Kullback–Leibler divergence to guide $Attn_{m}$ towards $Attn_{sup}$. Additionally, we apply a valid mask $M_v$ to avoid misleading $Attn_{m}$ to attend regions masked out in $\hat{x}_t$:
\begin{equation}
    \begin{split}
L_{attn} &= \frac{1}{L}\sum_{i=1}^L \text{KL}(Attn_{sup}^i, Attn_{m}^i) \odot M_v \\
&M_v[j,:] =\left\{
\begin{aligned}
M  & , & M'^j=1, \\
1-M  & , & M'^j=0.
\end{aligned}
\right.
    \end{split}
  \label{eq:attn_loss}
\end{equation}
Finally, our learning objective $L$ includes the combined \firstcomponent~and the~\secondcomponent:
\begin{equation}
L=L_{pred} + \lambda_{attn} L_{attn}.
  \label{eq:losses}
\end{equation}
In inference, we only use self-attention modules for the target domain.
\begin{table*}
  \caption{\small Comparison with baselines on the GTAV-to-Cityscapes benchmark. The ${\rm mIoU}$ represents the average of individual mIoUs among all 19 categories between GTAV and Cityscapes. Arch. represents the architecture of the model. C represents convnet-based model while T represents Transformer.  The best results are highlighted in bold while the second best results are shown by underline. }
  \centering
  \resizebox{\textwidth}{!}{
  \begin{tabular}{ccccccccccccccccccccc|c}
    \toprule
    Method& Arch. &\rotatebox{60}{Road}&\rotatebox{60}{Sidewalk}&\rotatebox{60}{Building}&\rotatebox{60}{Wall}&\rotatebox{60}{Fence}&\rotatebox{60}{Pole}&\rotatebox{60}{Tr.Light}&\rotatebox{60}{Tr.Sign}&\rotatebox{60}{Veget.}&\rotatebox{60}{Terrain}&\rotatebox{60}{Sky}&\rotatebox{60}{Person}&\rotatebox{60}{Rider}&\rotatebox{60}{Car}&\rotatebox{60}{Truck}&\rotatebox{60}{Bus}&\rotatebox{60}{Train}&\rotatebox{60}{Motor.}&\rotatebox{60}{Bicycle}&mIoU \\
    \midrule
    Source only & C & 75.8 & 16.8 & 77.2 & 12.5 & 21.0 & 25.5 & 30.1 & 20.1 & 81.3 & 24.6 & 70.3 & 53.8 & 26.4 & 49.9 & 17.2 & 25.9 & 6.5 & 25.3 & 36.0 & 36.6 \\
    CBST~\cite{ZouYuKuWa18}& C & 91.8 & 53.5 & 80.5 & 32.7 & 21.0 & 34.0 & 28.9 & 20.4 & 83.9 & 34.2 & 80.9 & 53.1 & 24.0 & 82.7 & 30.3 & 35.9 & 16.0 & 25.9 & 42.8 & 45.9 \\
    DACS~\cite{TranhedenOlPiSv21}& C & 89.9 & 39.7 & 87.9 & 30.7 & 39.5 & 38.5 & 46.4 & 52.8 & 88.0 & 44.0 & 88.8 & 67.2 & 35.8 & 84.5 & 45.7 & 50.2 & 0.0 & 27.3 & 34.0 & 52.1 \\
    CorDA~\cite{corda}& C & 94.7 & 63.1 & 87.6 & 30.7 & 40.6 & 40.2 & 47.8 & 51.6 & 87.6 & 47.0 & 89.7 & 66.7 & 35.9 & 90.2 & 48.9 & 57.5 & 0.0 & 39.8 & 56.0 & 56.6 \\
    ProDA~\cite{proda}& C & 87.8 & 56.0 & 79.7 & 46.3 & 44.8 & 45.6 & 53.5 & 53.5 & 88.6 & 45.2 & 82.1 & 70.7 & 39.2 & 88.8 & 45.5 & 59.4 & 1.0 & 48.9 & 56.4 & 57.5 \\
    \hline
    Source only & T & 71.5 & 18.0 & 84.2 & 34.4 & 30.9 & 33.4 & 44.3 & 23.5 & 87.4 & 41.3 & 86.6 & 64.0 & 22.5 & 88.3 & 44.5 & 39.1 & 2.3 & 35.2 & 31.6 & 46.5 \\
    DAFormer~\cite{HoyerDaGo22}& T & \underline{95.7} & \underline{70.2} & \underline{89.4} & \underline{53.5} & \underline{48.1} & \underline{49.6} & \textbf{55.8} & \underline{59.4} & \textbf{89.9} & \underline{47.9} & \textbf{92.5} & \textbf{72.2} & \underline{44.7} & \underline{92.3} & \underline{74.5} & \underline{78.2} & \underline{65.1} & \textbf{55.9} & \underline{61.8} & \underline{68.3} \\
    Ours & T & \textbf{96.5} & \textbf{73.9} & \textbf{89.5} & \textbf{56.8} & \textbf{48.9} & \textbf{50.7} & \textbf{55.8} & \textbf{63.3} & \textbf{89.9} & \textbf{49.1} & \underline{91.2} & \textbf{72.2} & \textbf{45.4} & \textbf{92.7} & \textbf{78.3} & \textbf{82.9} & \textbf{67.5} & \underline{55.2} & \textbf{63.4} & \textbf{69.6} \\
    \hline
    Oracle & T & 98.0 & 84.2 & 92.6 & 59.4 & 59.7 & 61.9 & 66.7 & 76.6 & 92.5 & 66.4 & 94.9 & 79.6 & 60.7 & 94.6 & 84.0 & 88.6 & 81.2 & 63.2 & 75.0 & 77.9 \\
    \bottomrule
  \end{tabular}
  }
  \label{tab:GTA_experiments}
\end{table*}

\section{Experiments}
In this chapter, we demonstrate experimental results under different settings in order to thoroughly verify the effectiveness and reliability of our proposed method. We introduce detailed experiment protocols in Sec.~\ref{sec:exp_prot} and compare our method with other existing baselines on different benchmarks in Sec.~\ref{sec:comp}. We also analyze the contribution from each of our vital model components in ablation studies in Sec.~\ref{sec:abl}, and further discuss the effects regarding our cross-domain attention in Sec.~\ref{sec:pert}. 
We present qualitative results by showing predictions obtained with our method in Sec.~\ref{sec:quali}. Finally, we verify our model's robustness when different hyper-parameter values are used, and its generalizability to different benchmarks in Sec.~\ref{sec:sensitivity}.

\begin{table*}
  \caption{\small Comparison with baselines on the Synthia-to-Cityscapes benchmark. The ${\rm mIoU}$ represents the average of individual mIoUs among all 16 categories between Synthia and Cityscapes.
  Arch. represents the architecture of the model. C represents convnet-based model while T represents Transformer.
  The best results are highlighted in bold while the second best results are shown by underline. 
  }
  \centering
  \resizebox{\textwidth}{!}{
  \begin{tabular}{cccccccccccccccccc|c}
    \toprule
    Method&Arch.&\rotatebox{60}{Road}&\rotatebox{60}{Sidewalk}&\rotatebox{60}{Building}&\rotatebox{60}{Wall}&\rotatebox{60}{Fence}&\rotatebox{60}{Pole}&\rotatebox{60}{Tr.Light}&\rotatebox{60}{Tr.Sign}&\rotatebox{60}{Veget.}&\rotatebox{60}{Sky}&\rotatebox{60}{Person}&\rotatebox{60}{Rider}&\rotatebox{60}{Car}&\rotatebox{60}{Bus}&\rotatebox{60}{Motor.}&\rotatebox{60}{Bicycle}&mIoU \\
    \midrule
    Source only & C &64.3 & 21.3 & 73.1 & 2.4 & 1.1 & 31.4 & 7.0 & 27.7 & 63.1 & 67.6 & 42.2 & 19.9 & 73.1 & 15.3 & 10.5 & 38.9 & 34.9 \\
    CBST~\cite{ZouYuKuWa18}& C &68.0 & 29.9 & 76.3 & 10.8 & 1.4 & 33.9 & 22.8 & 29.5 & 77.6 & 78.3 & 60.6 & 28.3 & 81.6 & 23.5 & 18.8 & 39.8 & 42.6 \\
    DACS~\cite{TranhedenOlPiSv21}&C & 80.6 & 25.1 & 81.9 & 21.5 & 2.9 & 37.2 & 22.7 & 24.0 & 83.7 & \underline{90.8} & 67.6 & 38.3 & 82.9 & 38.9 & 28.5 & 47.6 & 48.3 \\
    CorDA~\cite{corda}&C & \textbf{93.3} & \textbf{61.6} & 85.3 & 19.6 & 5.1 & 37.8 & 36.6 & 42.8 & 84.9 & 90.4 & 69.7 & 41.8 & 85.6 & 38.4 & 32.6 & 53.9 & 55.0 \\
    ProDA~\cite{proda}&C & \underline{87.8} & \underline{45.7} & 84.6 & 37.1 & 0.6 & 44.0 & 54.6 & 37.0 & \textbf{88.1} & 84.4 & \underline{74.2} & 24.3 & \textbf{88.2} & 51.1 & 40.5 & 45.6 & 55.5 \\
    \hline
    Source only&T & 51.5 & 20.3 & 79.2 & 19.3 & 1.8 & 40.9 & 29.9 & 22.7 & 79.1 & 82.4 & 63.0 & 24.9 & 75.8 & 33.7 & 18.9 & 24.9 & 35.2 \\
    DAFormer~\cite{HoyerDaGo22}&T & 84.5 & 40.7 & \textbf{88.4} & \textbf{41.5} & \underline{6.5} & \underline{50.0} & \underline{55.0} & \textbf{54.6} & \underline{86.0} & 89.8 & 73.2 & \textbf{48.2} & \underline{87.2} & \underline{53.2} & \underline{53.9} & \textbf{61.7} & \underline{60.9} \\
    Ours &T & 83.7 & 42.9 & \underline{87.4} & \underline{39.8} & \textbf{7.5} & \textbf{50.7} & \textbf{55.7} & \underline{53.5} & 85.9 & \textbf{90.9} & \textbf{74.5} & \underline{47.2} & 86.0  & \textbf{60.2} & \textbf{57.8} & \underline{60.8} & \textbf{61.5} \\
    \hline
    Oracle & T & 98.0 & 84.2 & 92.6 & 59.4 & 59.7 & 61.9 & 66.7 & 76.6 & 92.5 & 94.9 & 79.6 & 60.7 & 94.6 & 88.6 & 63.2 & 75.0 & 78.0 \\
    \bottomrule
  \end{tabular}
  }

    \vspace{-2mm}
  \label{tab:SYN_experiments}
\end{table*}

\begin{table*}
  \caption{\small Comparison with baselines on the Cityscapes-to-ACDC benchmark. The results are obtained on the held-out test set of ACDC whose annotation is not accessible. Arch. represents the architecture of the model. C represents convnet-based model while T represents Transformer.  The best results are highlighted in bold while the second best results are shown by underline. 
}
  \centering
  \vspace{-2mm}
  \resizebox{\textwidth}{!}{
  \begin{tabular}{ccccccccccccccccccccc|c}
    \toprule
    Method&Arch.&\rotatebox{60}{Road}&\rotatebox{60}{Sidewalk}&\rotatebox{60}{Building}&\rotatebox{60}{Wall}&\rotatebox{60}{Fence}&\rotatebox{60}{Pole}&\rotatebox{60}{Tr.Light}&\rotatebox{60}{Tr.Sign}&\rotatebox{60}{Veget.}&\rotatebox{60}{Terrain}&\rotatebox{60}{Sky}&\rotatebox{60}{Person}&\rotatebox{60}{Rider}&\rotatebox{60}{Car}&\rotatebox{60}{Truck}&\rotatebox{60}{Bus}&\rotatebox{60}{Train}&\rotatebox{60}{Motor.}&\rotatebox{60}{Bicycle}&mIoU \\
    \midrule
    ADVENT~\cite{VuJaBuCoPe19}& C &72.9 & 14.3 & 40.5 & 16.6 & 21.2 & 9.3 & 17.4 & 21.3 & 63.8 &23.8 & 18.3 & 32.6 & 19.5 & 69.5 & 36.2 & 34.5 & 46.2 & 26.9 & 36.1 & 32.7 \\
    BDL~\cite{LiYuVa19}& C &56.0 & 32.5 & 68.1 & 20.1 & 17.4 & 15.8 & 30.2 & 28.7 & 59.9 & 25.3 & 37.7 & 28.7 & 25.5 & 70.2 & 39.6 & 40.5 & 52.7 & 29.2 & 38.4 & 37.7 \\
    CLAN~\cite{LuoZhGuYuYa19}& C &\textbf{79.1} & 29.5 & 45.9 & 18.1 & 21.3 & 22.1 & 35.3 & 40.7 & 67.4 & 29.4 & 32.8 & 42.7 & 18.5 & 73.6 & 42.0 & 31.6 & 55.7 & 25.4 & 30.7 & 39.0 \\
    FDA~\cite{YangSo20}& C &\underline{74.6} & \textbf{73.2} & 70.1 & \textbf{63.3} & \textbf{59.0} & \textbf{54.7} & \underline{52.3} & 47.0 & 44.9 & 44.8 & 43.3 & 39.5 & \textbf{34.7} & 29.5 & 28.6 & 28.5 & 28.3 & 28.2 & 24.8 & 45.7 \\
    MGCDA~\cite{SakaridisDaGo22}& C &73.4 & 28.7 & 69.9 & 19.3 & 26.3 & 36.8 & \textbf{53.0} & 53.3 & \textbf{75.4} & 32.0 & \textbf{84.6} & 51.0 & 26.1 & 77.6 & 43.2 & 45.9 & 53.9 & 32.7 & 41.5 & 48.7 \\
    \hline
    DAFormer~\cite{HoyerDaGo22}& T &58.4 & \underline{51.3} & \underline{84.0} & 42.7 & \underline{35.1} & \underline{50.7} & 30.0 & \textbf{57.0} & \underline{74.8} & \underline{52.8} & 51.3 & \underline{58.3} & 32.6 & \underline{82.7} & \underline{58.3} & \underline{54.9} & \underline{82.4} & \textbf{44.1} & \underline{50.7} & \underline{55.4} \\
    Ours & T &57.6 & 43.7 & \textbf{85.1} & \underline{43.5} & 33.9 & 50.1 & 42.9 & \underline{53.9} & 72.8 & \textbf{52.9} & \underline{52.2} & \textbf{59.4} & \textbf{34.7} & \textbf{83.6} & \textbf{60.4} & \textbf{68.7} & \textbf{84.3} & \underline{41.4} & \textbf{53.0} & \textbf{56.5} \\
    \bottomrule
  \end{tabular}
  }
  \label{tab:ACDC_experiments}
\end{table*}

\subsection{Experiment Protocols} \label{sec:exp_prot}
We implemented using the MMSegmentation library. We take DAformer as our backbone model and followed the same configurations such as learning rate, etc. We adopted MiT-B5~\cite{XieWaYuAnAlLu21} pretrained on ImageNet as our encoder, and decoder from DAFormer. We also followed the optimization policies and chose AdamW~\cite{LoshchilovHu19} as the optimizer with the learning rate of $6\times 10^{-5}$ for the encoder, $6\times 10^{-4}$ for the decoder with weight decay of 0.01 and batch size of 2. DACS~\cite{TranhedenOlPiSv21}, Learning rate warmup, rare class sampling (RCS), thing-class ImageNet feature distance (FD) and other tricks applied in DAFormer were also kept with the original configurations including hyper-parameters. 

\noindent\textbf{Dataset.} 
GTAV~\cite{RichterViRoKo16} is a synthetic street scene dataset rendered by a game engine with pixel-level annotations. It includes 24,966 images with resolution $1914 \times 1052$. 
Synthia~\cite{RosSeMaVaLo16} is another synthetic street scene dataset containing 9,400 images with the corresponding annotation with a resolution $1280 \times 760$. 
Cityscapes~\cite{CordtsOmRaReEnBeFrRoSc16} is a real-world street scene dataset that includes 2,975 training images and 500 held-out images for evaluation with resolution of $2048 \times 1024$. 
ACDC~\cite{SakaridisDaGo21} is also a real-world street scene dataset sharing 19 common categories with Cityscapes under adverse conditions including foggy, nighttime, rainy, and snowy scenarios. There are 1,600 training images, 406 validation images, and, more importantly, 2,000 held-out testing images with resolution $1920 \times 1080$.
We follow our baselines to resize GTAV images to resolution $1280 \times 720$, Synthia images to resolution $1280 \times 760$, Cityscapes images to resolution $1024 \times 512$, and ACDC images to $960 \times 540$ and crop images to resolution $512 \times 512$ during the training.

\noindent\textbf{Baselines and Metric.} 
We choose representative and state-of-the-art UDA CNN-based baselines including CBST~\cite{ZouYuKuWa18}, ADVENT~\cite{VuJaBuCoPe19}, DACS~\cite{TranhedenOlPiSv21}, CorDA~\cite{corda}, MGCDA~\cite{SakaridisDaGo22}, ProDA~\cite{proda}, and transformer-based baseline DAformer. Following~\cite{HoyerDaGo22}, we report mean Intersection over Union (mIoU) as our evaluation metrics averaged over three different random seeds.

\subsection{Comparison with Existing Methods} \label{sec:comp}
\noindent\textbf{Comparative Evaluation.} We compare our method with other existing state-of-the-art methods on the different three benchmarks. Besides the commonly-adopted synthetic-to-real benchmarks GTAV-to-Cityscapes in Tab.~\ref{tab:GTA_experiments} and Synthia-to-Cityscapes in Tab.~\ref{tab:SYN_experiments}, on which the performance on the Cityscapes' validation set is reported, we also verified our method on the normal-to-adverse adaptation benchmark Cityscapes-to-ACDC in Tab.~\ref{tab:ACDC_experiments} and report the results on its test set. 
We especially advocate that it is important to evaluate and report the performance on a held-out test set, as most existing works in UDA semantic segmentation literature focus only on the performance on the validation set of Cityscapes if these models are tuned on the same validation set. Without testing on the held-out test set, we cannot ensure whether model hyper-parameters are generalizable to other adaptation benchmarks without overfitting.
During the experiments, we set $\lambda_{attn}$ to 1, which is the only hyper-parameter of our method, according to the performance on the validation set on ACDC, and applied the same value to the experiments on all other benchmarks. Further details about the hyper-parameter selection will be discussed in Sec.~\ref{sec:sensitivity}.

\noindent\textbf{Evaluation Results.} As presented in all three tables, the introduction of Transformer already brings in improvements over the convnet-based model when only source domain data are available. DAFormer significantly boosts the performance by introducing adequate modifications and training strategies for Transformer under the UDA segmentation settings and outperforms other baselines by a noticeable margin, which reveals the outstanding prediction capacity of Transformer. Our method outperforms DAFormer on all three benchmarks, which suggests that our method further mitigates the domain gap with attention-level and output-level consistency learning.  On average, our method provides 1.3 percent point (pp) over the state-of-the-art performance on GTAV-to-Cityscapes, 0.6 pp on Synthia-to-Cityscapes, and 1.1 pp on the test set of Cityscapes-to-ACDC. This shows not only the effectiveness of our proposed method in aligning domain gaps under different scenarios but also its reliability and generalizability to be expanded to other types of domain shift.

\subsection{Ablation Study} \label{sec:abl}
We verify the effectiveness of our \firstcomponent~and the~\secondcomponent, and analyze their contribution of each component of our model in Tab.~\ref{table:ablation}. We also report a standard deviation over three runs.
S$\&$T represents the performance of DAFormer. We report the performance of each component, the combination of these components, and our final model. 

\begin{table}[t]
\caption{\small Ablation studies on the GTAV-to-Cityscapes benchmark. S\&T represents the performance of DAFormer, T2S and S2T represent the $L_{t2s}$ and $L_{s2t}$ in Eq.~\ref{eq:cda_losses}. Attn represent $L_{Attn}$ in Eq.~\ref{eq:attn_loss}. Grad. does not use gradient stopping on the query vectors (By default, we use gradient stopping on the query vectors).}
\centering
\setlength{\tabcolsep}{8.5pt}
\begin{tabular}{ c c c c c c } 
 \toprule
 & S\&T & T2S & S2T & Attn & mIoU  \\
 \midrule
  1 & $\checkmark$ &  &  &  & $68.3\pm0.7$ \\
  2 & $\checkmark$ & $\checkmark$ &  &  & $68.1\pm0.8$ \\
  3 & $\checkmark$ &  & $\checkmark$ &  &  $68.5\pm0.2$  \\
  4 & $\checkmark$& $\checkmark$ & $\checkmark$&  & $69.1\pm0.6$ \\
  5 & $\checkmark$&  &  & $\checkmark$ & $68.6\pm0.2$ \\
  6 & $\checkmark$& Grad. & Grad. & $\checkmark$ & $69.1\pm0.8$ \\
  7 & $\checkmark$& $\checkmark$ & $\checkmark$& $\checkmark$ & $69.6\pm0.5$ \\
 \bottomrule
\end{tabular}
\label{table:ablation}
\end{table}
\begin{figure*}[t]
  \centering
   \includegraphics[width=\linewidth]{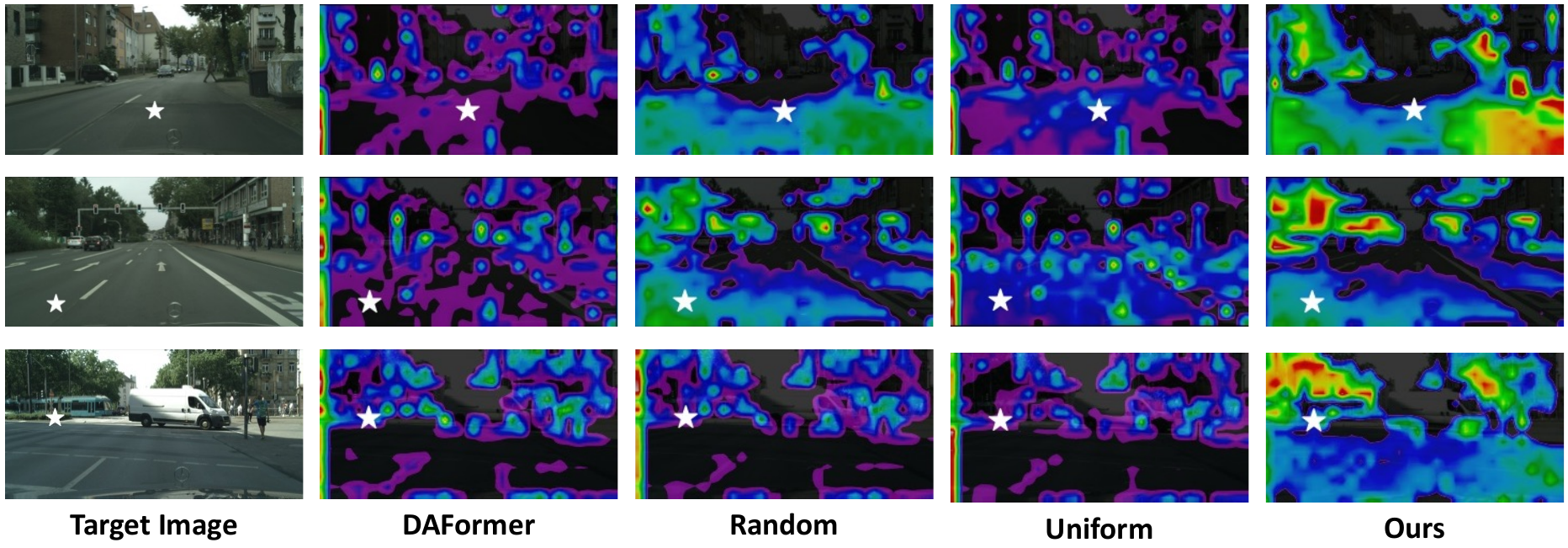}
   \caption{\small Attention map visualization for the query pixel $\largestar$ from different models trained with a different type of regularization methods. }
   \label{fig:pert}
\end{figure*}

Based on the results, we observe that individually adding the target-to-source (T2S) or source-to-target (S2T) prediction consistency loss provides small improvements over the baseline, while combining these two consistency regularizations helps better and surpasses the baseline by 0.8 pp (from row 1 to 4) of improvement in mIoU, while \secondcomponent provides 0.3 pp (from row 1 to 5). 
Comparing the improvements brought by the~\firstcomponent (\ie, row 1 to 4 and row 5 to 7)  \vs ~\secondcomponent (\ie, row 1 to 5 and row 4 to 7), we observe that the \firstcomponent~brings in relatively more improvements to the mean value of mIoU (0.8 pp from row 1 to 4 and 1.0 pp from row 5 to 7), while the~\secondcomponent~contributes mainly on decreasing the standard deviation (-0.5 pp from row 1 to 5 and -0.1 pp from row 4 to 7).
This could imply the different effects of our attention and output level alignment: the former helps better in narrowing down the random factors and strengthening the robustness of the training while the latter plays a more important role in directly boosting the accuracy. Finally, when we combine the \firstcomponent~and the~\secondcomponent, our full model achieves the best performance. 

Additionally we test our query gradient stopping and compare the performance in row 6 and 7. We see that by stopping the gradient flowing back through the query in the cross-domain branches, we can improve the performance by 0.5 pp with even lower computation complexity. We conjecture that the improvement comes from the less aggressive alignment in the cross-domain branches, as the cross-domain attention map itself can be very noisy and possibly provide higher gradient signals to the wrong regions or pixels due to the domain discrepancy. 



\subsection{A Study of Perturbations on Attention Maps} \label{sec:pert}
We explained in earlier chapters that the effectiveness of our cross-domain attention could partially come from its perturbation to the self-attention in Transformer and therefore encourage more robust learning.
We additionally try several perturbations in the self-attention in DAFormer.  
We first try to simply perturb the attention maps in both the source and the target branches in DAFormer by adding synthetic Gaussian noise (denoted by $+$ Random). 
To better simulate attention maps, we generate noise from the normal distribution in a smaller resolution, upsample it via bilinear interpolation to the actual size of an attention map to smoothen it, and then calculate the attention score through a softmax layer. 
Finally we take the average of the original attention maps with the synthetic noise to simulate the perturbation on original self-attention.
Meanwhile we also try another recent work~\cite{NamKiHeHaOhOh2} that advocates introducing uniform attention maps during the training phase (denoted by $+$ Uniform). Distracted by the denser uniform attention, the model will be forced to exploit more informative and crucial signals from the self-attention, and the model can therefore benefit from the concentrated attention to achieve better performance with barely extra computational cost. Technically, it requires only a context broadcasting (CB) module after the self-attention module to manually inject uniformity into self-attention. 
We also add the CB module on both the source and the target branches of the DAFormer model with original configurations. The experiment results for both the random attention and the uniform attention are reported in Tab.~\ref{tab:pert}, and the visualization of their attention maps are compared in Fig.~\ref{fig:pert}.


\begin{table}[t]
\centering
\caption{\small Comparison with different regularization methods on attention map with the DAFormer backbone. GTAV indicates the results obtained on GTAV-to-Cityscapes while Synthia indicates those on Synthia-to-Cityscapes.  }
\centering
\resizebox{0.45\textwidth}{!}{\begin{tabular}{ c | c c  c c } 
 \toprule
    Benchmark & DAFormer & $+$ Random & $+$ Uniform & $+$ Ours \\
    \midrule
    GTAV  & $68.3$ & $67.3 $ &  $66.9 $ & $69.6$ \\
    Synthia & $60.9$ &  $60.4 $ & $59.1 $ & $61.5$  \\
 \bottomrule
\end{tabular}}
\label{tab:pert}
\end{table}

\begin{figure*}[t]
  \centering
   \includegraphics[width=\linewidth]{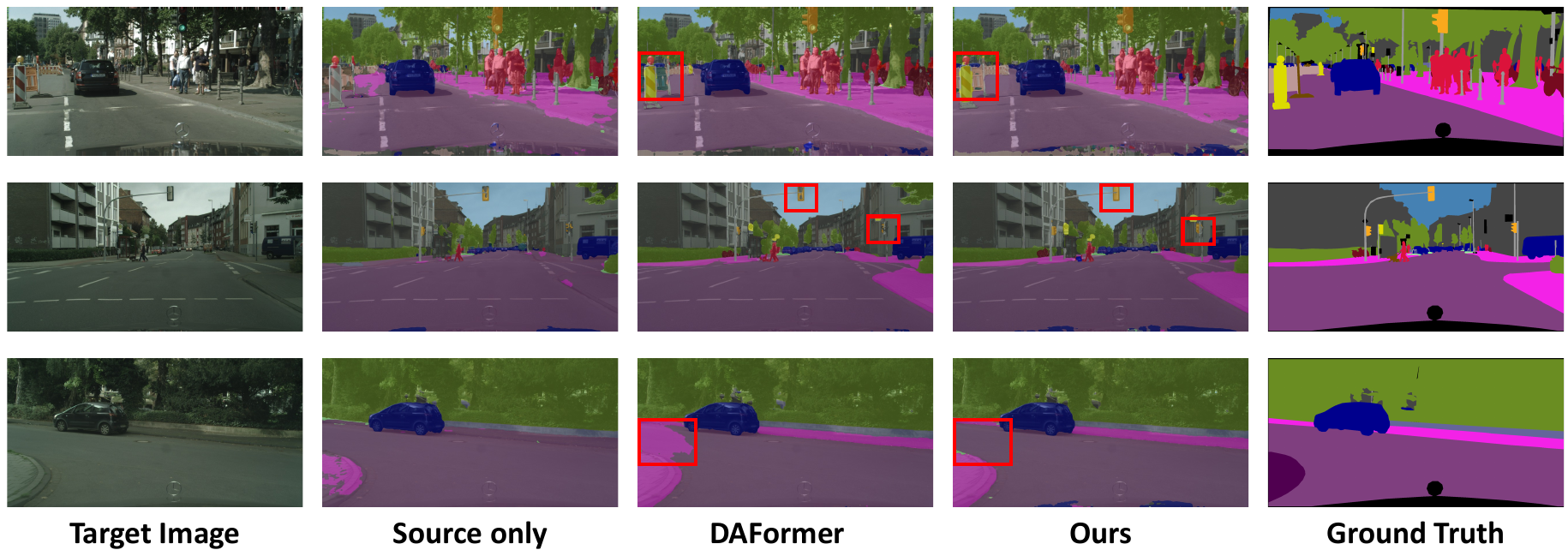}
   \caption{\small Qualitative results of our method versus the source-only baseline and the DAFormer. }
   \label{fig:quali}
\end{figure*}

We can observe that both the aforementioned perturbation methods fail to provide improvement compared to the performance of DAFormer on GTAV-to-Cityscapes (68.3\%) and Synthia-to-Cityscapes (60.9\%). From the Fig.~\ref{fig:pert}, The random perturbation alleviates spurious attention in some cases (the first and second rows). However, uniform attention does not improve attention maps. The models trained with these synthetic perturbations still suffer from the attention-level domain gaps similar to DAFormer.
The quantitative results in Tab.~\ref{tab:pert} also show random and uniform attention even hurt the performance of DAFormer. 
We believe that the reason is because those perturbations do not help attention-level alignment, and it is not enough to improve the performance under UDA settings by naively perturbing the self-attention without addressing the domain gap issues.
In comparison, our method shares query vectors across domains in the attention module and thus manages to satisfy both adaptation and perturbation.  It therefore better addresses the domain discrepancy on different levels, as shown qualitatively in Fig.~\ref{fig:pert} and quantitatively in Sec.~\ref{sec:comp}.

\subsection{Qualitative Results} \label{sec:quali}
We present qualitative results in Fig.~\ref{fig:quali}. In these samples, we observe that DAFormer and ours obtain higher quality predictions over the source-only model. At the same time, we also observe that our method generally obtains finer and better predictions than that of DAFormer (\eg, the fence in the first row, the traffic lights in the second row, and the road in the third row highlighted by red boxes).

\subsection{Sensitivity Analysis} \label{sec:sensitivity}
\begin{figure}[t]
\centering
\vspace{-2mm}
    \begin{subfigure}[t]{0.15\textwidth}
        \includegraphics[width=0.99\linewidth]{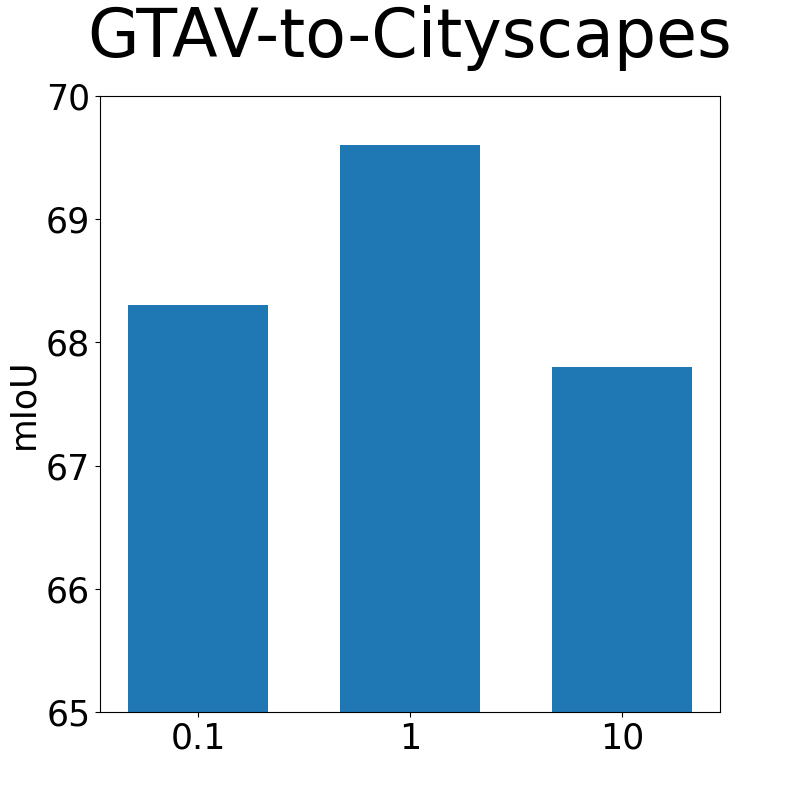}  
    \end{subfigure}
    \begin{subfigure}[t]{0.15\textwidth}
        \includegraphics[width=0.99\linewidth]{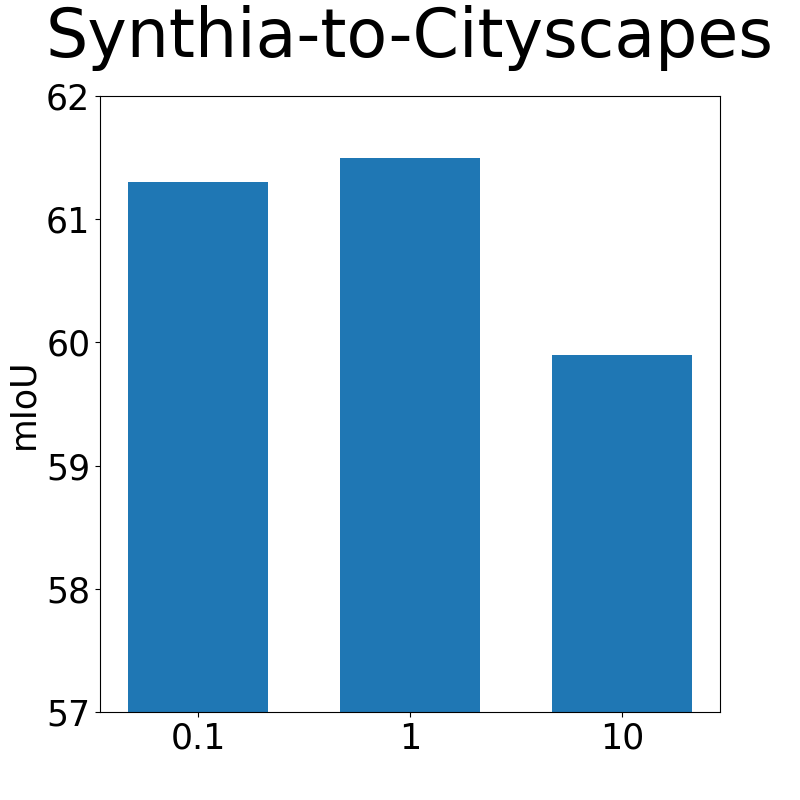}
    \end{subfigure}
    \begin{subfigure}[t]{0.15\textwidth}
        \includegraphics[width=0.99\linewidth]{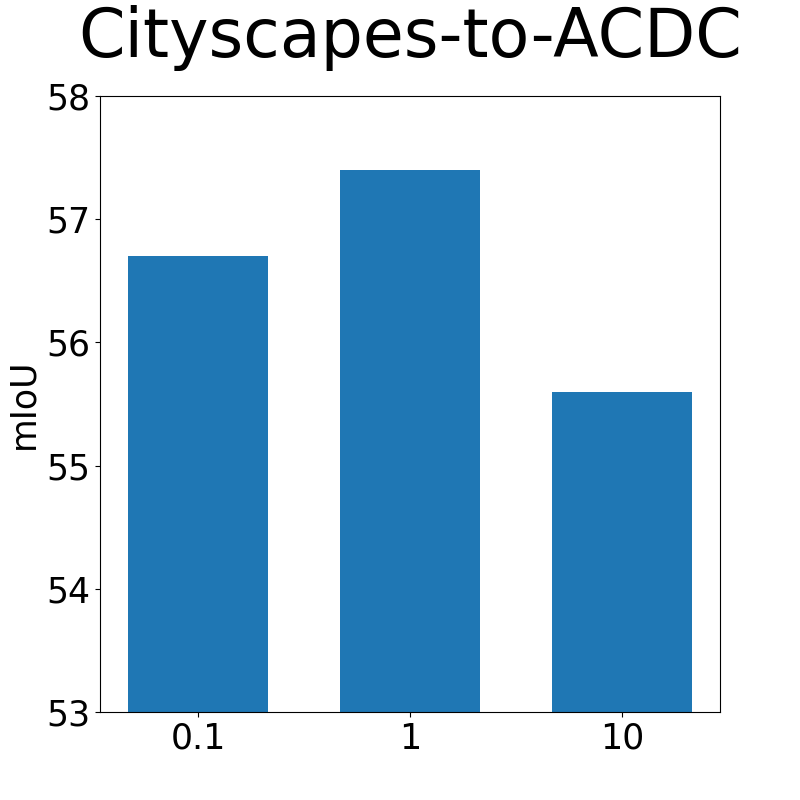}
    \end{subfigure}
    \caption{\small Sensitivity analysis on $\lambda_{attn}$. Our method shows stable performance over hyper-parameters on different benchmarks.}
    \label{fig:sen}
\end{figure}

As we introduced in Sec.~\ref{sec:comp}, the verification of the generalizability of UDA segmentation methods is regularly ignored in many existing works, and we cannot exclude the possibility that their methods are overtuned on a few specific adaptation tasks or target datasets. To this end, we tested the sensitivity of our method to its hyper-parameter, \ie, $\lambda_{attn}$.
Specifically, we ran experiments with $\lambda_{attn}=$0.1, 1 and 10 on GTAV-to-Cityscapes, Synthia-to-Cityscapes and Cityscapes-to-ACDC. We report validation mIoU on Cityscapes and ACDC, in Fig.~\ref{fig:sen}. 

Based on the results, we can see that our current setting of $\lambda_{attn}=1$ stably achieves the best performance on all three sets. When the attention weight is lower, it does not provide strong enough help to attention-level alignment and affects the overall performance. On the other hand, the accuracy plummets when the attention weight goes too high, as the consistency learning starts suppressing other learning objectives.
More importantly, we observe a similar tendency of performance with respect to $\lambda_{attn}$, which indicates that our model is generalizable, rather than dependent on a specific group of hyperparameters or benchmarks.

\section{Conclusion}
Despite the significant improvement brought by the recent introduction of Transformers in UDA semantic segmentation, Transformers still suffers from noticeable attention-level domain discrepancy. 
In this work, we leverage consistency between the predictions from self-attention and cross-domain attention and address both attention-level and output-level domain shifts. 
Further assisted by our~\secondcomponent, our model is more robust to features from different levels on both domains and therefore improves the performance on the target domain.
Extensive experiments as well as ablation studies verify the effectiveness, reliability of our method, and the generalizability under different types of domain shift.

\clearpage
\section{Acknowledgements}
This work has been partially supported by ONR MURI grant N00014-19-1-2571 associated with AUSMURIB000001.

{\small
\bibliographystyle{ieee_fullname}
\bibliography{egbib}
}

\end{document}